\documentclass[letterpaper, 10 pt, conference]{ieeeconf}  

\usepackage{graphicx}
\graphicspath{{images/}}
\usepackage{amsmath}
\usepackage{subcaption}
\usepackage[thinlines]{easytable}
\usepackage{cite}
\usepackage{float}

\IEEEoverridecommandlockouts                              

\graphicspath{{images/}}

\usepackage{color}
\usepackage{booktabs}
\usepackage{textcomp}
\usepackage{microtype} 
\usepackage{hyperref}
\usepackage{ccicons}  

\usepackage{xspace}

\usepackage{tabularx}
\usepackage[labelfont=bf,font=footnotesize]{caption}
\usepackage{amssymb}
\usepackage{multirow}
\usepackage{verbatim}
\usepackage{booktabs}

\renewcommand{\baselinestretch}{0.97}

\makeatletter
\DeclareRobustCommand\onedot{\futurelet\@let@token\@onedot}
\def\@onedot{\ifx\@let@token.\else.\null\fi\xspace}

\makeatother
\usepackage{lipsum}

\usepackage[dvipsnames]{xcolor}     
\usepackage{amsfonts}               
\usepackage{mathrsfs}
\usepackage{algorithm}             
\usepackage[noend]{algpseudocode}  

\usepackage{listings}              
\lstdefinestyle{custompy}{
  belowcaptionskip=1\baselineskip,
  breaklines=true,
  xleftmargin=\parindent,
  language=Python,
  showstringspaces=false,
  basicstyle=\footnotesize\ttfamily,
  keywordstyle=\bfseries\color{green!40!black},
  commentstyle=\itshape\color{purple!40!black},
  identifierstyle=\color{blue},
  stringstyle=\color{orange},
}

\long\def\xspace{\mathcal{X}}

\title{\LARGE \bf
Parts-Based Articulated Object Localization in Clutter \\Using Belief Propagation
}

 \author{Jana Pavlasek \and Stanley Lewis \and Karthik Desingh \and Odest Chadwicke Jenkins%
 \thanks{J. Pavlasek, S. Lewis, K. Desingh and O. C. Jenkins are with the Department of Electrical Engineering and Computer Science, Robotics Institute, University of Michigan, Ann Arbor, MI, USA 48109 \newline
         {\tt\small \{pavlasek, stanlew, kdesingh, ocj\}@umich.edu}}
\thanks{This work was supported in part by NSF award IIS-1638060.}
 }

\begin{document}

\maketitle
\thispagestyle{plain}
\pagestyle{plain}

\begin{abstract}
Robots working in human environments must be able to perceive and act on challenging objects with articulations, such as a pile of tools. Articulated objects increase the dimensionality of the pose estimation problem, and partial observations under clutter create additional challenges. To address this problem, we present a generative-discriminative parts-based recognition and localization method for articulated objects in clutter. We formulate the problem of articulated object pose estimation as a Markov Random Field (MRF). Hidden nodes in this MRF express the pose of the object parts, and edges express the articulation constraints between parts. Localization is performed within the MRF using an efficient belief propagation method. The method is informed by both part segmentation heatmaps over the observation, generated by a neural network, and the articulation constraints between object parts. Our generative-discriminative approach allows the proposed method to function in cluttered environments by inferring the pose of occluded parts using hypotheses from the visible parts. We demonstrate the efficacy of our methods in a tabletop environment for recognizing and localizing hand tools in uncluttered and cluttered configurations.
The project webpage is available at: \href{http://janapavlasek.com/projects/tool-parts/}{\it janapavlasek.com/projects/tool-parts}.
\end{abstract}

\section{Introduction} \label{sec:intro}
Robot assistants operating in real-world environments should be capable of performing maintenance and repair tasks.
Going beyond pick-and-place actions, we aim to enable robots to use the diversity of objects it might encounter.
The ability to use commercial, off-the-shelf hand tools is critical for robots to perform tasks in unstructured, everyday environments.
In order to accomplish this, robots must be able to identify and localize tools in an arbitrary cluttered scene to plan appropriate actions toward performing a task.

\begin{figure}[t!]
  \centering
  \includegraphics[width=0.98\linewidth]{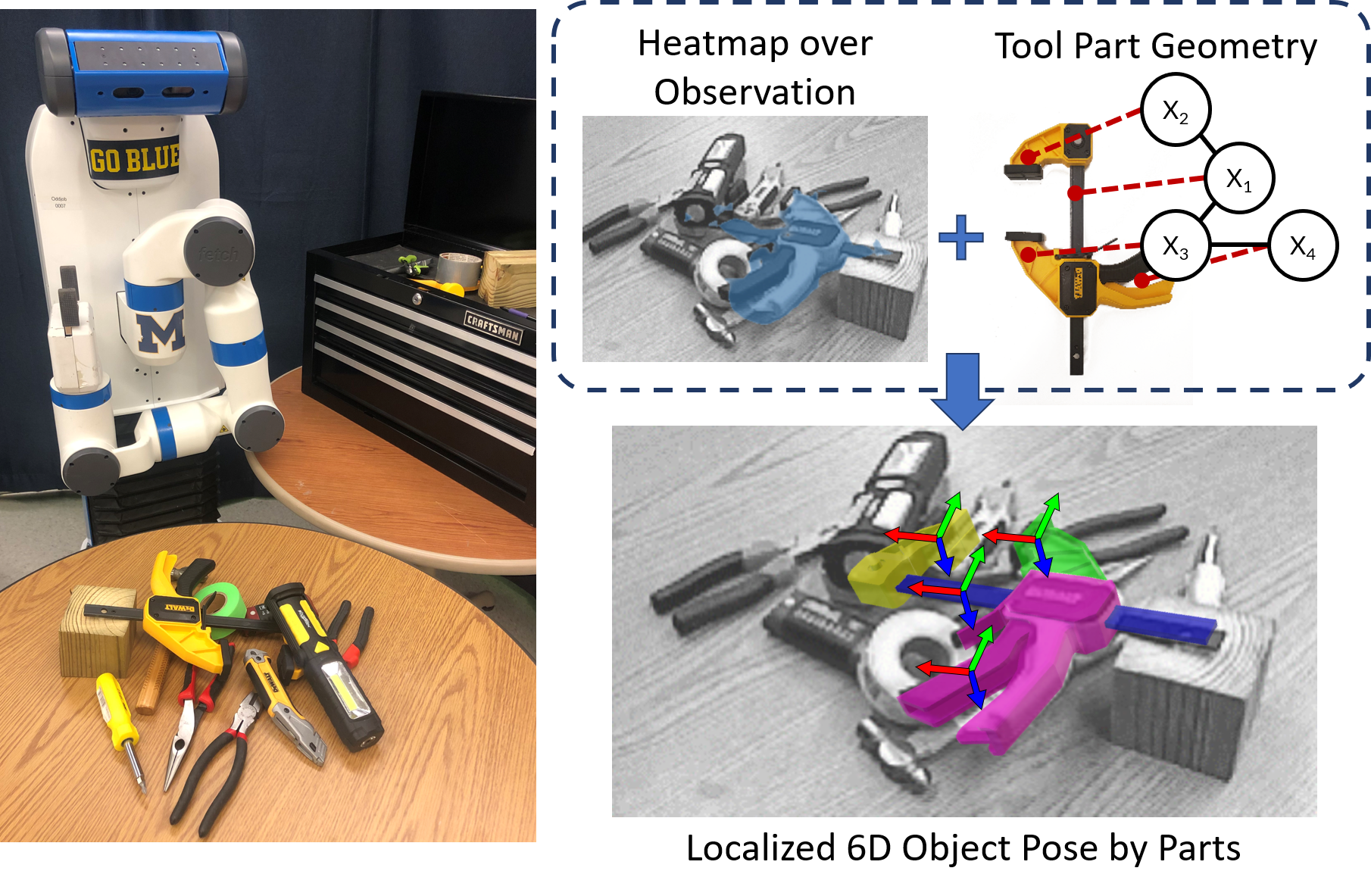}
  \caption{\label{fig:motivation} The ProgressLab Fetch robot perceiving a cluttered scene with tools (articulated objects): The top right image shows the pixelwise heatmap over the RGB observation for the clamp object, which misses the top part of the clamp as it is occluded by another object. Our parts-based localization method is able to use this partially informed heatmap along with the clamp geometry to localize the 6D pose of the entire object (bottom right image).}
\end{figure}

Recognizing hand tools and localizing their pose remains challenging in common human environments. These challenges arise from uncertainty caused by physical clutter and the high-dimensionality of the space of poses multiple objects in contact may occupy.
Many hand tools are articulated, adding complexity to the localization problem by introducing additional degrees-of-freedom.
Figure~\ref{fig:motivation} shows one example of hand tools in a cluttered scene that could be typical in a work area.

State-of-the-art object and pose recognition methods have been proposed that estimate the six degree-of-freedom (6D) pose of objects using convolutional neural networks (CNNs)~\cite{xiang2018posecnn, tremblay2018corl:dope}. Other methods have accomplished pose estimation using probabilistic inference \cite{sui2015axiomatic, deng2019poserbpf}.
However, localizing \textit{articulated} objects remains a challenge for these methods due to both the added degrees of freedom that arise from articulations and occlusions due to clutter.
Parts-based representations~\cite{Felzenszwalb} have the potential to achieve higher levels of robustness under these conditions than whole-object based approaches.  For such methods to be suitable for robot manipulation tasks, they must be able to localize 6D pose with reasonable computational efficiency.
We suggest that generative inference methods, if made more computationally efficient, offer compelling and complementary benefits to modern deep learning. Additionally, a parts-based representation can provide information about the affordances of an object, because robot actions are typically applied to the object parts.

In this paper, we present a method for recognition and localization of articulated objects in clutter suited to robotic manipulation of object affordances.
We formulate the problem of articulated object pose estimation as a Markov Random Field (MRF), representing the 6D poses of each rigid object part and the articulation constraints between them.
We propose a method to perform inference over the MRF based on message passing. We are inspired by work by Desingh et al.~\cite{Desingheaaw4523}, in which parts-based articulated object localization is facilitated by combining information from both the observation as well as the compatibility with neighbouring parts within the inference process.
Our method is informed jointly by a learned likelihood modelled by a CNN, as well as by the known articulation constraints between each component part.
We assume known object mesh models and kinematic constraints in the form of a Unified Robot Description Format (URDF) file, a standard geometrical object representation in the field of robotics.

By employing generative inference to integrate both data-driven techniques and domain knowledge about the object models, we leverage the speed and representational abilities of deep CNNs, while retaining the ability to reconcile noisy results and provide structure and context to the estimate.
Methods we present emphasize novel synthesis of
(1) efficient discriminative-generative inference via nonparametric belief propagation for pose estimation of articulated objects, and
(2) a  learned  part-based likelihood to evaluate hypotheses of articulated object pose against RGB observations.
We present results using a custom dataset made up of commercial, off-the-shelf hand tools with robot observations containing varying levels of clutter.

\section{Related Work} \label{sec:related_work}
Pose estimation has received considerable attention in robotics.
Here, we discuss related work that focuses on rigid body, parts-based, and articulated object pose estimation.

\subsection{Rigid Body Pose Estimation}
Methods that tackle the problem of rigid body pose estimation include geometry-based registration approaches~\cite{besl1992method}, generative approaches~\cite{sui2015axiomatic, desingh2016physically}, approaches combining discriminative and generative methods~\cite{narayanan2017deliberative, sui2017sum, chen2019grip, zeng2018semantic, mitash2018robust, deng2019poserbpf}, and end-to-end learning approaches~\cite{xiang2018posecnn, tremblay2018corl:dope}. Here, we focus on the discriminative-generative methods and end-to-end learning methods that are most relevant to this work.

Combining the discriminative power of feature-based methods with generative inference has been successful under challenging conditions such as background and foreground clutter~\cite{narayanan2017deliberative,  mitash2018robust, deng2019poserbpf}, adversarial environment conditions~\cite{chen2019grip}, and uncertainty due to robot actions~\cite{sui2017sum, zeng2018semantic}. We are inspired by the success of the above approaches in taking advantage of the speed of discriminative methods to perform analysis and synthesis based generative inference.

Xiang et al. propose an end-to-end network for estimating 6D pose from RGB images~\cite{xiang2018posecnn}. This work was further extended to use synthetic data generation and augmentation techniques to improve performance~\cite{tremblay2018corl:dope}. Wang et al.~\cite{wang2019densefusion} propose an end-to-end network that uses depth information along with RGB information.
These methods rely significantly on the textured appearance of objects.
More importantly, the state representation used in these methods assume rigidity. Our attempts to adapt these methods for articulated objects required considerably more training data and computation time.
In addition, estimates from the end-to-end methods can be noisy, especially in challenging cluttered scenarios.
We believe estimates from these methods will be a good prior to help generative methods recover under challenging scenarios.
Hence, in this work, we learn a likelihood function over the observation to inform the generative inference.

\subsection{Parts-Based Pose Estimation}

Understanding objects in terms of their parts paves the way to meaningful and purposeful action execution, such as tool-use.
Parts-based representations have been proposed to aid scene understanding and action execution~\cite{Felzenszwalb, felzenszwalb2005pictorial,  xiang2012estimating}, and have recently garnered attention within the robotics and perception communities~\cite{Mo2018, Lu2018}. Parts-based localization has led to research in recognizing objects and their articulated parts \cite{Yi2018}. Parts-based perception for objects in human environments is often limited to recognition and classification tasks. Parts-based pose estimation is often considered for human body pose~\cite{sigal2004tracking} and hand pose~\cite{sudderth2004visual} estimation with fixed graphical models. Here, we propose a general framework for estimating pose of articulated objects, such as hand tools, that includes parts with fixed transforms as constraints.

\subsection{Articulated Pose Estimation and Tracking}
Probabilistic inference is a popular technique in robot perception for articulated body tracking~\cite{cifuentes2016probabilistic, schmidt2014dart, schmidt2015depth}, where filtering-based approaches alongside novel observation models have been proposed. These tracking frameworks are either initialized to the ground truth poses of objects, or applied to robot manipulators, where the inference is informed by joint encoder readings. In this work, we aim to perform pose estimation of multiple articulated objects using a single RGB-D frame with weak initialization from pixel-wise segmentations.

Interactive perception~\cite{interactiveperception} for articulated object estimation~\cite{hausman2015artic} has been a problem of interest in the robotics community. Various works~\cite{martin14online, sturm11prob, sturm13book}, propose methods for estimating kinematic models from demonstration of manipulation or articulation examples. We instead focus on using known kinematic models to estimate the objects in challenging cluttered environments.

Li et al.~\cite{li2019category} explore category-level localization of articulated bodies in a point cloud, however their method does not consider clutter and occlusions from the environment. Michel et al.~\cite{michel2015pose} perform one-shot pose estimation of articulated bodies using 3D correspondences with optimization over hypotheses. Desingh et al. consider pose estimation of articulated objects in cluttered scenarios using efficient belief propagation~\cite{Desingheaaw4523}, but do not consider RGB information. All of these approaches consider large, primarily planar objects that cover significant portion of the observation as opposed to the small objects in clutter in this work.

Li et al.~\cite{li2016hierarchical} developed techniques to handle the challenges of hand tools and small objects with no articulation, however the techniques proposed require multi-viewpoint information, as opposed to the single image approach that we propose.

\section{Problem Statement} \label{sec:problem}
\begin{figure}
 \centering
 \includegraphics[width=0.45\linewidth]{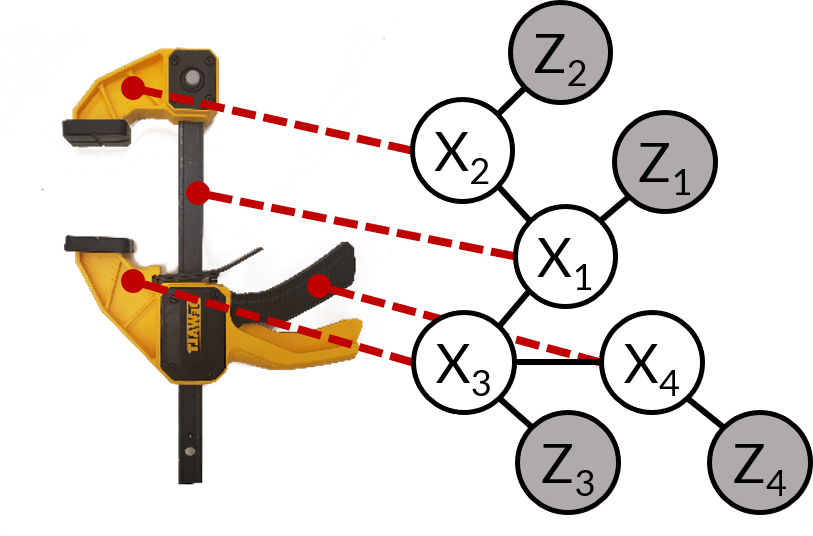}
 \caption{\label{fig:mrf} The Markov Random Field representation of the clamp object. The clamp is broken up into four parts to fully represent both its affordances and articulations. The hidden nodes (white) represent the pose of each part, $X_i$. The observed nodes (grey) correspond to the sensor observation.}
\end{figure}

\begin{figure*}
    \centering
    \captionsetup[subfigure]{labelformat=empty}
    \includegraphics[width=0.78\textwidth]{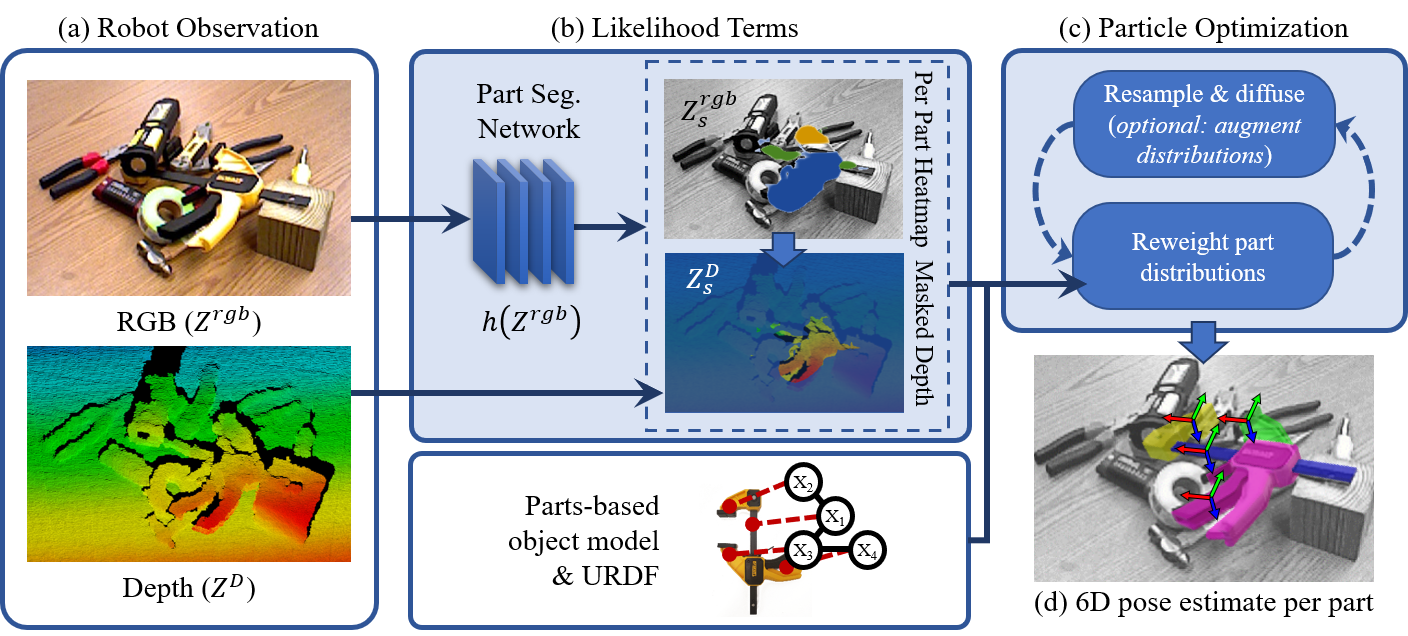}
 \caption{\footnotesize The inference pipeline. (a) The robot observes a scene as an RGB-D image, $\mathcal{Z}=(Z^{rgb}, Z^D)$. (b) The RGB image is passed through a trained part segmentation network, $h(Z^{rgb})$, that generates a pixel-wise heatmap for the $P_k$ parts of an object class of interest, $\{Z_s^{rgb}\}_{s=1}^{P_k}$ (in this example, the clamp, which has one fully occluded part). The heatmaps are used to generate masked depth images, $\{Z_s^{D}\}_{s=1}^{P_k}$ (c) The inference is initialized with part poses using these heatmaps and the depth image. Hypotheses are iteratively reweighed using Equation \ref{eq:reweight_mp}, and resampled with importance sampling. (d) The inference process generates an estimate of the 6D pose of each part. (Best viewed in color).}
 \label{fig:pipeline}
\end{figure*}

Given a scene containing objects $O$, such that $\{O_k\}_{k=1}^K$ is the set of $K$ relevant objects, we wish to localize each object $O_k$.
The state of an object $O_k$ is represented by the set of part poses $\mathcal{X}=\{X_i\}_{i=1}^{P_k}$, where $X_s$ is the 6D pose of an articulating rigid part $s$ of $O_k$, with $P_k$ parts. Each object $O_k$ in the scene is estimated independently.

This estimation problem is formulated as a Markov Random Field (MRF).
Let $G=(V, E)$ denote an undirected graph with nodes $V$ and edges $E$. An example MRF is illustrated in Figure \ref{fig:mrf}.
The joint probability of the graph $G$ is expressed as:
\begin{equation}
\label{eq:jointprob}
    p(\mathcal{X}, \mathcal{Z}) \propto \prod_{(s,t)\in E}\psi_{s,t}(X_s, X_t) \prod_{s \in V}\phi_{s}(X_s, Z_s)
\end{equation}
where $\mathcal{X}$ denotes the hidden state variables to be inferred and $\mathcal{Z}$ denotes the observed sensor information in the form of an RGB-D image. The function $\psi_{s,t}$ is the \textit{pairwise potential}, describing the correspondence between part poses based on the articulation constraints, and $\phi_{s}$ is the \textit{unary potential}, describing the correspondence of a part pose $X_s$ with its observation $Z_s$. The problem of pose estimation of an articulated model $O_k$ is interpreted as the problem of estimating the marginal distribution of each part pose, called the belief, $bel(X_s)$.

In addition to the sensor data, the articulation constraints and 3D geometry of the object, in the form of a Unified Robot Description Format (URDF), and the 3D mesh models of the objects are provided as inputs. We assume that the object articulations are produced by either fixed, prismatic or revolute joints. We consider scenes which contain only one instance of an object. In Section~\ref{sec:method}, our proposed inference mechanism is detailed, along with a description of our modelling of the potentials in Equation~\ref{eq:jointprob}.

\section{Methodology} \label{sec:method}

Belief propagation via iterative message passing is a common approach to infer hidden variables while maximizing the joint probability of a graphical model \cite{Desingheaaw4523, sudderth2004visual}.
We adopt the sum-product iterative message passing approach to perform inference~\cite{wainwright2008graphical}, where messages are passed between hidden variables until their beliefs converge. A message, denoted by $m_{ts}^{n}(X_s)$, can be considered as the belief of the receiving node $s$ as informed by its neighbor $t$ at iteration $n$. An approximation of the message, denoted by $\hat{m}_{ts}^{n}(X_s)$, is computed using the incoming messages to $t$:
\begin{equation}
\label{eq:max_msg}
    \hat{m}_{ts}^{n}(X_s) = \sum_{X_t \in \mathbb{X}_t} \phi_t(X_t, Z_t) \psi_{s,t}(X_s, X_t)\prod_{u\in \rho(t)\setminus s}\hat{m}_{ut}^{n-1}(X_t)
\end{equation}
where $\rho(t)$ denotes neighboring nodes of $t$, and $\mathbb{X}_t$ denotes the particle set of node $t$.

The marginal belief of a hidden node is a product of all the incoming messages weighted by the node's unary potential:
\begin{equation}
\label{eq:marginal_belief}
    bel_s^{n}(X_s) \propto \phi_s(X_s, Z_s) \prod_{t\in \rho(s)} \hat{m}_{ts}^{n}(X_s)
\end{equation}
Our particle optimization algorithm aims to approximate the joint probability of the MRF, as in Equation \ref{eq:jointprob}, by maintaining the marginal belief, as in Equation \ref{eq:marginal_belief} for each object part. The belief of a rigid part pose, $bel_s(X_s)$, is represented nonparametrically as a set of $N$ weighted particles $\mathbb{X}_s = \{X_s^{(i)}, w_s^{(i)}\}_{i=1}^N$.

Section \ref{sec:method_particle_filter} describes the message passing algorithm. Section \ref{sec:method_unary} describes how the function $\phi_s(X_s, Z_s)$ is represented, and Section \ref{sec:method_pairwise} describes the function $\psi_{s,t}(X_s, X_t)$.

\subsection{Belief Propagation via Message Passing} \label{sec:method_particle_filter}

Our method adopts the traditional reweight and resample paradigm for particle refinement methods.
The particles are first \textit{reweighted} using an approximated sum-product message.
The particles are then \textit{resampled} using importance sampling based on the calculated weights.

The high-dimensional nature of the estimation problem and the cluttered settings with similar parts and partial observations make the inference prone to convergence to local minima. To mitigate this problem while computing messages, we can optionally add an augmentation step before the reweight step to accommodate different proposals. The augmentation technique is adapted from Pacheco et al.~\cite{pacheco2014preserving} and is discussed in Section \ref{sec:method_augmentation}.

The overall system is summarized in Figure~\ref{fig:pipeline}.

\subsubsection{Reweighting and Resampling steps}
Each particle $X_s^{(i)} \in \mathbb{X}_s$ is reweighted as follows:
\begin{equation}
\label{eq:reweight_mp}
    w_s^{(i)} = \phi_s(X_s^{(i)}, Z_s) \prod_{t\in \rho(s)} \hat{m}_{ts}^{n}(X_s^{(i)})
\end{equation}
where $\hat{m}_{ts}^{n}(X_s^{(i)})$ is the sum-product message:
\begin{equation}
\label{eq:local_max_msg}
    \hat{m}_{ts}^{n}(X_s^{(i)}) = \sum_{X_t^{(j)} \in \mathbb{X}_t} \psi_{s,t}(X_s^{(i)}, X_t^{(j)})\phi_{t}(X_t^{(j)}, Z_t)
\end{equation}
which only takes into account the immediate neighbors of the node. Since the number of parts in each object is small, this approximation has negligible effect in practice and saves computation time.
For numerical stability, the log-likelihoods are used in practice.
The weights are normalized and then the particles are resampled using importance sampling.
The object pose estimate is made by selecting the maximum likelihood estimates (MLE) from each of the marginal beliefs.

\subsection{Unary Likelihood} \label{sec:method_unary}

The unary potential represents the compatibility of each pose hypothesis with the RGB-D observation, $\mathcal{Z}$. The RGB and depth portions of the observation, $Z^{rgb}$ and $Z^D$, are treated as independent such that the unary likelihood is:
\begin{equation} \label{eq:unary_rgbd}
    \phi_s(X_s, Z_s) = \phi_{s}^\mathrm{rgb}(X_s, Z_s^{rgb}) \phi_s^{D}(X_s, Z_s^D)
\end{equation}
where $\phi_s^{D}$ and $\phi_{s}^{\mathrm{rgb}}$ are the likelihoods with respect to depth and RGB parts of the observations.

\subsubsection{RGB Unary Likelihood}
The RGB portion of the unary likelihood makes use of the Dilated ResNets architecture \cite{Yu2017}. This architecture maintains a high dimensional feature space which is beneficial for semantic segmentation tasks.

The CNN outputs a pixelwise score for each object part class $s$. We apply a sigmoid function so the final scores lie between zero and one. This constitutes a learned heatmap $Z_s^{rgb}=h_s(Z^{rgb})$ over an RGB observation $Z^{rgb}$, where $h$ is the Dilated ResNets model trained on parts and $h_s(\cdot)$ is the output indexed at class $s$.
For each particle hypothesis $X_s$, we generate a mask $\mathcal{M}_s$ over the image for the object part at the hypothesis pose. We transform the mesh model of the part to pose $X_s$ and use the camera parameters to obtain a corresponding binary mask in image space. We represent the likelihood of a particle $X_s$ over the heatmap $Z_s^{rgb}$ using the Jaccard index~\cite{Moulton2018Jaccard}, commonly called the Intersection over Union (IoU), between the heatmap and the rendered mask:
\begin{equation}
    \phi_{s}^\mathrm{rgb}(X_s, Z_s^{rgb}) = \frac{|\mathcal{M}_s \cap Z_s^{rgb}|}{|\mathcal{M}_s| + |Z_s^{rgb}| - |\mathcal{M}_s \cap Z_s^{rgb}|}
\end{equation}

The CNN is trained using the analagous intersection over union (IoU) loss, and as such, $\phi_{s}^\mathrm{rgb}(X_s)$ represents a learned likelihood function over the image.

\subsubsection{Depth Unary Likelihood}
For a given part $s$, depth observation $Z_{s}^D$ is generated using a threshold over the heatmap $Z_s^{rgb}$ to mask the depth image $Z^D$. For a particle $X_s$, $\phi^D(X_s, Z_{s}^D)$ is the exponential of the negative average pixelwise error between $Z_{s}^D$ and the mesh model of part $s$, rendered at pose $X_s$.
The error is only evaluated over areas in which the two depth images overlap. If there is no overlap between the masked observation and the hypothesis, we assign a maximum error instead, which is a chosen constant.

\subsection{Pairwise Likelihood} \label{sec:method_pairwise}

The pairwise likelihood between neighbouring particles $\psi_{t,s}(X_t, X_s)$ measures how compatible $X_s$ is with respect to $X_t$. If $X_s$ falls within the joint limits of $s$ with respect to $t$ at pose $X_t$, then $\psi_{t,s}(X_t, X_s)=1$. Otherwise, the likelihood is the exponential of the negative error between $X_s$ and the nearest joint limit. We refer to~\cite{Desingheaaw4523} for further details.

\subsection{Particle Augmentation} \label{sec:method_augmentation}

At each node $s$, the particle set $\mathbb{X}_s$ can be augmented by drawing particles from various proposal distributions. Given $N$ particles in $\mathbb{X}_s$, Gaussian noise is first added to the current particles, then the distribution is augmented to $\mathbb{X}_s^{prop}= \mathbb{X}_s \cup \mathbb{X}_s^{aug}$, where $\mathbb{X}_s^{aug}$ represents the particles generated from the augmentation procedure $q$. The set $\mathbb{X}_s^{prop}$ contains $\alpha N$ particles, where $\alpha > 1$.
Various proposals $q_s^{pair}$, $q_s^{unary}$, and $q_s^{rand}$, as described below can be used to augment the particle set. This optional variant is evaluated and discussed in the results section.

\textbf{\textit{Pairwise:}} The pairwise proposal distribution $q_s^{pair}(X_s) \propto \psi_{s, t}(X_s, \tilde{X}_t)$ is conditioned on a sample $\tilde{X}_t$, drawn from neighboring node $t$. Using the known geometric relationship between nodes $t$ and $s$, a compatible proposal for node $s$, $\tilde{X}_s$, is generated from $\tilde{X}_t$.

\textbf{\textit{Unary:}} The unary proposal distribution $q_s^{unary}(X_s) \propto \phi_s(X_s, Z_s)$ draws samples based on the unary potential $\phi_s$.

\textbf{\textit{Random:}} The random proposal distribution $q_s^{rand}(X_s) \propto \mathcal{N}(X_s, \Sigma)$ draws additional noisy samples. This can be used to avoid the belief falling into a local minima due to the high dimensionality of the orientation space, and to account for mirror symmetry in some objects.

\section{Experiments} \label{sec:results}

We evaluate our methods for articulated object localization in uncluttered and cluttered scenes. We run experiments on each component of our method and provide an analysis of their effects. These results provide quantitative and qualitative evidence of the accuracy and practicality of our methods.

We test on 20 uncluttered and 17 cluttered test scenes, unseen in the training data. We localize 196 total object instances in these scenes. We do not include results on objects which are severely or fully occluded such that there is no clear observation of any part. We remove 19 objects which fall into this category. An example of such a case is shown in the highly cluttered scene in Figure \ref{fig:compelling_rgb}, where the flashlight (behind the hammer) and lineman's pliers (behind the clamp) are almost entirely occluded.

\subsection{Dataset \& Training}

\begin{figure}
  \centering
  \includegraphics[width=0.8\linewidth]{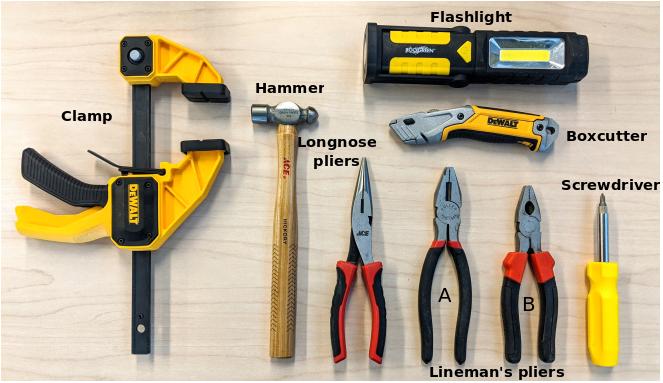}
  \caption{\label{fig:dataset} Objects in our custom dataset. The clamp has one prismatic joint. The three sets of pliers each have one revolute joint. The other four objects are treated as rigid. The objects are separated into parts based on the presence of both affordances and articulations.}
\end{figure}

Our custom dataset consists of hand tools with eight distinct tool instances: hammer, clamp, boxcutter, flashlight, screwdriver, longnose pliers, and two instances of lineman's pliers (see Figure \ref{fig:dataset}). We collect videos of both cluttered and uncluttered scenes using the Fetch Mobile Manipulator's onboard Primesense Carmine 1.09 sensor. The articulated hand tools span the full range of possible articulations in the data. Semantic masks and 6D poses for the objects are labelled using Label Fusion \cite{marion2018label}, which generates annotations for each video once the first scene is manually labelled. Semantic part masks and part poses are calculated using the object URDFs. The pixels in the images which do not correspond to a tool part are given class label ``background.'' After downsampling to remove adjacent frames in the videos, the dataset contains $\sim 6k$ RGB-D images of $640\times 480$ pixel resolution.

We train the Dilated ResNets \textit{DRN-D-22} architecture~\cite{Yu2017} to perform semantic segmentation on 90\% of the dataset, and reserve 10\% for validation. We further augment the training images with random crops, flips, and rotations. We increase the training set size by applying two transforms per image. The backbone is pre-trained on ImageNet, and the last layers are finetuned on our dataset. We employ the Intersection over Union (IoU) loss with an Adam optimizer.
We train for 10 epochs on a RTX 2080 Max-Q GPU.

\subsection{Implementation Details}

Our implementation performs efficient unary potential computation on the GPU,
to evaluate the heatmap from DRN and to generate binary masks and depth images for pose hypotheses.
The current implementation is vectorized and processes all object parts in $\sim 0.5-2s$ for one iteration with 300 particles. The computation time could be further reduced with more efficient implementation.

The $x$ and $y$ locations of particle poses are initialized randomly in areas corresponding to high heat pixels of the heatmap over the RGB observation. The $z$-axis is initialized to the corresponding depth in the observed depth image. The initial orientations are uniformly distributed. For completely occluded parts which do not appear on the segmentation mask, we generate compatible poses from the neighbour initializations.

\subsection{Evaluation Metric}
For evaluation, we use the average point matching error proposed by Hinterstoisser et al.~\cite{Hinterstoisser2013pose}, which measures the average point pairwise distance between the rigid object model's point cloud in the ground truth and estimated poses:
\begin{equation} \label{eq:add}
m(P_{gt}, \hat{P}) = \frac{1}{\mathbb{N}}\sum_{(p_{gt}, \hat{p})\in (P_{gt}, \hat{P})} ||\hat{p} - p_{gt}||
\end{equation}
where $(p_{gt}, \hat{p}) \in (P_{gt}, \hat{P})$ are corresponding points in the ground truth and estimated point clouds respectively, each with $\mathbb{N}$ points in the rigid object model. We also report the symmetric point matching error, which measures the average pairwise distance between points in the estimated point cloud and the \textit{nearest} point in the ground truth point cloud:
\begin{equation} \label{eq:adds}
m_{\mathrm{sym}}(P_{gt}, \hat{P}) = \frac{1}{\mathbb{N}}\sum_{\hat{p}\in \hat{P}} \min_{p_{gt}\in P_{gt}}||\hat{p} - p_{gt}||
\end{equation}
The symmetric matching error represents the error in symmetric objects, such as the screwdriver, better by not penalizing estimates rotated around a degree of symmetry in the object. However, it tends to provide artificially low errors for incorrect estimates.

\begin{table*}
\renewcommand{\arraystretch}{1.3}
    \centering
    \caption{Average matching errors $m$ (cm) and symmetric matching errors $m_{\mathrm{sym}}$ (cm)
    }
    \begin{tabular}{r | c c | c c | c c |}
    \hline
        \textbf{Method} & \multicolumn{2}{c|}{\textbf{Cluttered}} & \multicolumn{2}{c|}{\textbf{Uncluttered}} & \multicolumn{2}{c|}{\textbf{Overall}}\\
         & $m$ & $m_{\mathrm{sym}}$ & $m$ & $m_{\mathrm{sym}}$ & $m$ & $m_{\mathrm{sym}}$ \\
    \hline
DRN+ICP & 6.59 $\pm$ 3.36 & 3.31 $\pm$ 2.69 & 6.38 $\pm$ 8.00 & 3.93 $\pm$ 10.84& 6.47 $\pm$ 6.32 & 3.65 $\pm$ 8.22\\
Parts-PF       & 6.03 $\pm$ 4.64 & 2.73 $\pm$ 4.04 & 4.55 $\pm$ 2.57 & 1.43 $\pm$ 1.33 & 5.23 $\pm$ 3.74 & 2.03 $\pm$ 2.98\\
MP+RGB       & 4.57 $\pm$ 3.96 & 2.83 $\pm$ 2.95 & 3.60 $\pm$ 2.75 & 2.04 $\pm$ 1.95 & 4.06 $\pm$ 3.40 & 2.41 $\pm$ 2.50\\
MP+RGB+ICP & 4.77 $\pm$ 3.89 & 3.03 $\pm$ 2.88 & 3.85 $\pm$ 2.65 & 2.29 $\pm$ 1.90 & 4.28 $\pm$ 3.32 & 2.64 $\pm$ 2.44\\
MP+RGB-D+Aug & 4.80 $\pm$ 3.80 & 2.90 $\pm$ 2.60 & 3.71 $\pm$ 2.55 & 2.18 $\pm$ 1.88 & 4.21 $\pm$ 3.23 & 2.51 $\pm$ 2.27\\
MP+RGB-D     & \textbf{3.58} $\pm$ 2.56 & \textbf{1.89} $\pm$ 1.84 & \textbf{2.65} $\pm$ 2.01 & \textbf{1.30} $\pm$ 1.21 & \textbf{3.08} $\pm$ 2.32 & \textbf{1.57} $\pm$ 1.56\\
    \hline
    \end{tabular}
    \label{tab:estimation_methods}
\end{table*}

\subsection{Baselines}

We implement two baselines, described below.

\textbf{\textit{Segmentation with ICP (DRN+ICP): }} We initialize the 3D position of the particle hypotheses using the depth image and segmentation mask generated by Dilated ResNets, with random orientations. We use Iterative Closest Point (ICP)~\cite{besl1992method} to find the transform from the initialized point cloud to the observed point cloud. ICP works best on local refinements, and is prone to failure when the initial orientation is incorrect. To accommodate for this failure, we generate $N$ proposal poses per part, perform ICP, and select the one with the best final fitness score as the estimate. In our experiments, $N=20$. A similar method is used by Wong et al.~\cite{wong2017segicp}.

\textbf{\textit{Part-based Particle Filter (Parts-PF): }} This baseline consists of independent particle filters at each tool part. We use the unary potential from Equation \ref{eq:unary_rgbd} to calculate the weights for each hypothesis, and use importance sampling to select particles at each iteration. We use 300 particles per part, and run for 85 iterations.

For both \textit{DRN+ICP} and \textit{Parts-PF} baselines, if a part is completely occluded in the image, a pose estimate cannot be generated from the segmentation. In such cases, we randomly select a neighboring part for which an estimate was made and use the object model to generate a corresponding pose for the occluded part. If the edge between the parts is articulated, a joint value is uniformly sampled within the joint limits.

\subsection{Parts-Based Pose Estimation}

\begin{figure}
  \centering
  \includegraphics[width=0.95\linewidth]{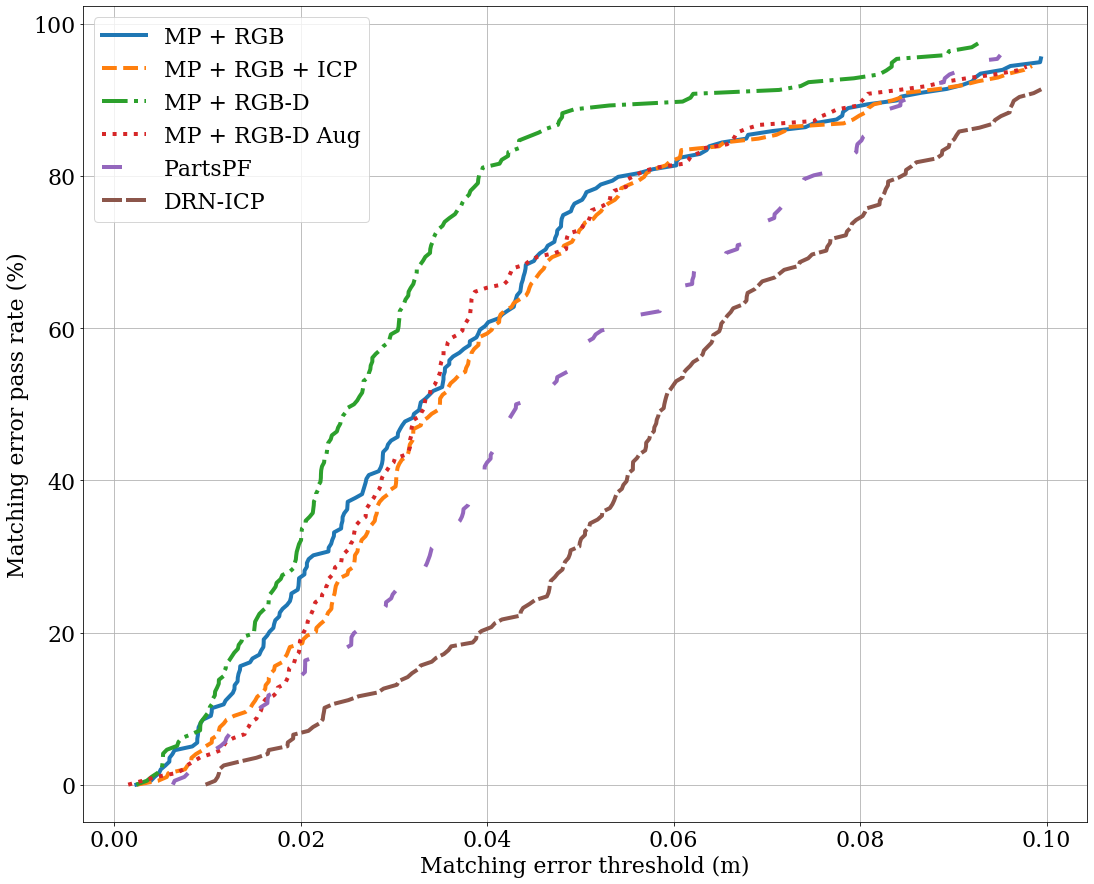}
  \caption{\label{fig:add_results} Average pairwise distance pass rate for each described method. All tools use the matching error described in Equation \ref{eq:add}, except the screwdriver, which due to its symmetrical nature, uses the symmetrical form ($m_{\mathrm{sym}}$).}
\end{figure}

To fully understand the performance of our method, we perform an ablation study over the components of the proposed method. We focus on three factors: the message passing (\emph{MP}), the use of \emph{RGB} only vs. the inclusion of depth (\textit{RGB-D}) in the unary potential, and the augmentation step (\emph{Aug}). We use 300 particles (before augmentation) for all experiments.
The choice of particles was observed qualitatively to achieve sufficient results in most cases. While representation of the underlying belief improves with more particles, computation becomes intractable for very large particle sets.
We run each method for 100 iterations, after which we observe little change in the estimate.

The results of each method are shown in Table \ref{tab:estimation_methods}. Figure \ref{fig:add_results} shows the results for all scenes. Message passing leads to superior results compared to the baselines, \emph{DRN-ICP} and \emph{Parts-PF}, which do not use message passing. Best performance is achieved by using the full RGB-D observation (\emph{MP-RGB-D}). Further description and analysis of each method is provided below.

\begin{figure}
  \centering
     \begin{subfigure}[b]{0.49\linewidth}
         \centering
         \includegraphics[width=0.9\textwidth]{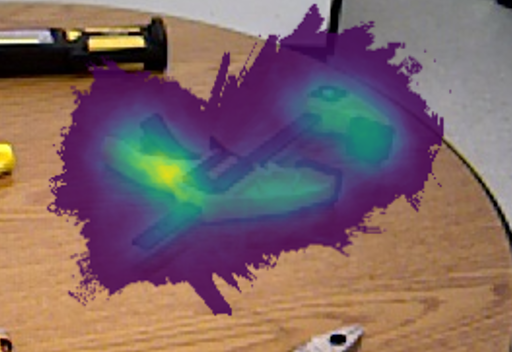}
         \caption{Belief for all parts from \emph{Parts-PF}.}
         \label{fig:fail_no_pair_bel}
     \end{subfigure}
     \hfill
     \begin{subfigure}[b]{0.49\linewidth}
         \centering
         \includegraphics[width=0.9\textwidth]{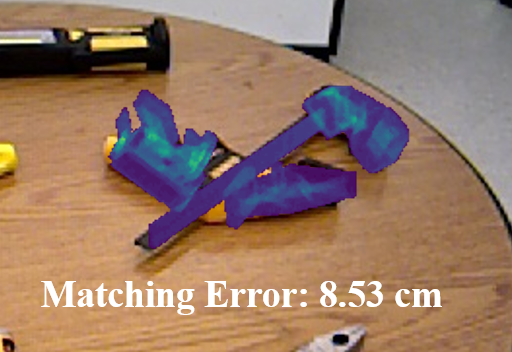}
         \caption{Estimate for each part from \emph{Parts-PF}.}
         \label{fig:fail_no_pair_est}
     \end{subfigure}
     \hfill
     \begin{subfigure}[b]{0.49\linewidth}
         \centering
         \includegraphics[width=0.9\textwidth]{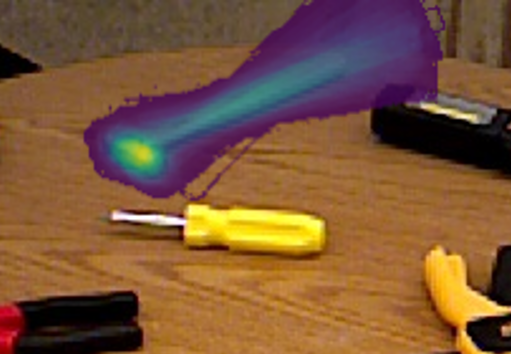}
         \caption{Belief for all parts from \emph{MP+RGB}}
         \label{fig:fail_rgb_img}
     \end{subfigure}
     \hfill
     \begin{subfigure}[b]{0.49\linewidth}
         \centering
         \includegraphics[width=0.9\textwidth]{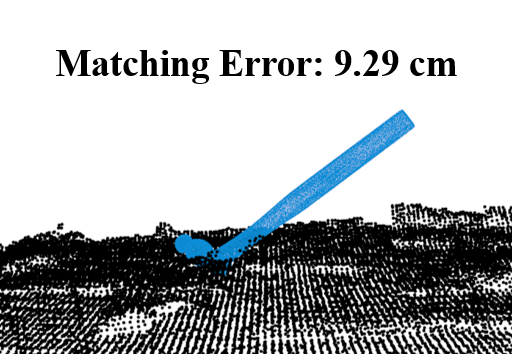}
         \caption{Estimate for each part from \emph{MP+RGB}}
         \label{fig:fail_rgb_depth}
     \end{subfigure}
    \caption{\label{fig:qualitative_failures} Common failure cases. The first row is a common failure of the baseline method, \emph{Parts-PF}, which does not use message passing. In (a) the parts are clustered in the correct positions in the image, but (b) shows that the part estimates are oriented incorrectly in an incompatible configuration. The second row is a common failure case in \emph{MP+RGB}. In (c), the particles have converged in the 2D image, but in (d), the hammer 6D pose is rotated about an axis which is not well represented in the 2D image plane.}
\end{figure}

\begin{figure*}
  \centering
  \includegraphics[width=\linewidth]{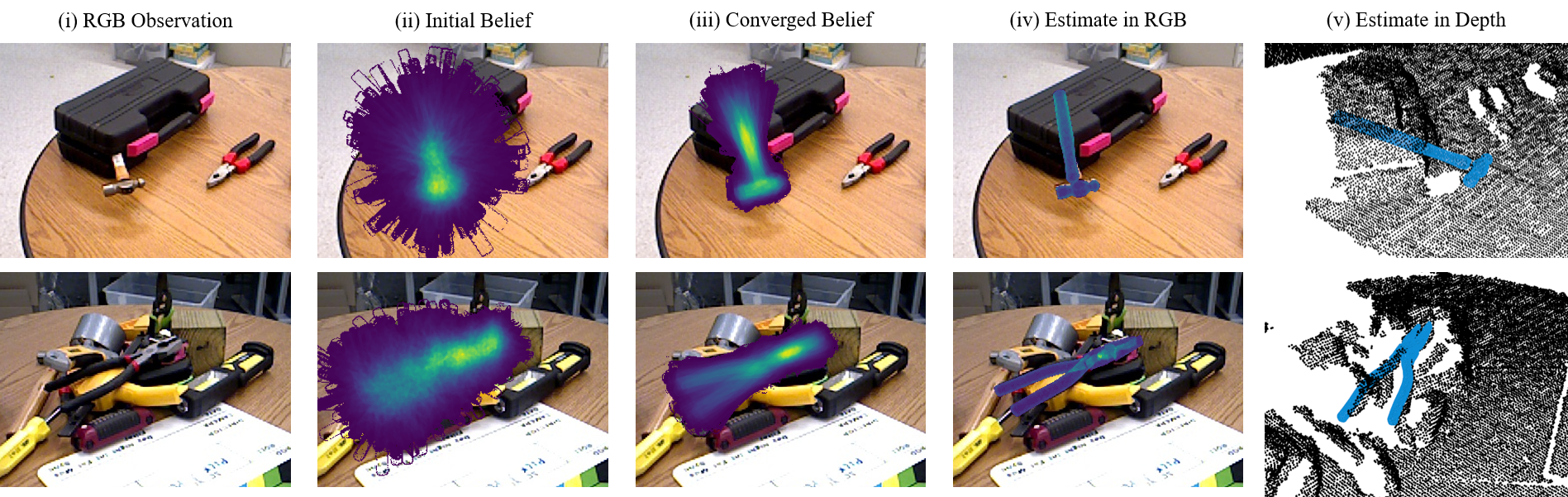}
  \caption{\label{fig:qualitative_method} Qualitative results for each stage of the \textit{MP+RGB-D} method on cluttered scenes. The method results in accurate pose estimates despite significant occlusions (top) and articulations (bottom).}
\end{figure*}

\begin{figure*}
     \centering
     \begin{subfigure}[b]{0.24\textwidth}
         \centering
         \includegraphics[width=0.95\textwidth]{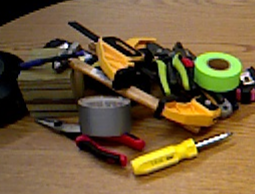}
         \caption{RGB input image}
         \label{fig:compelling_rgb}
     \end{subfigure}
     \hfill
     \begin{subfigure}[b]{0.24\textwidth}
         \centering
         \includegraphics[width=0.95\textwidth]{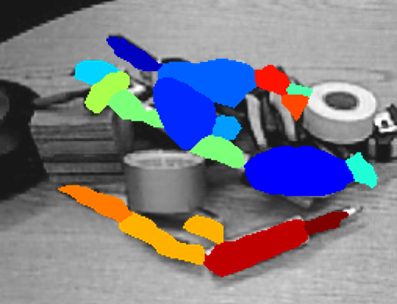}
         \caption{Segmentation mask from CNN}
         \label{fig:compelling_seg}
     \end{subfigure}
     \hfill
     \begin{subfigure}[b]{0.24\textwidth}
         \centering
         \includegraphics[width=0.95\textwidth]{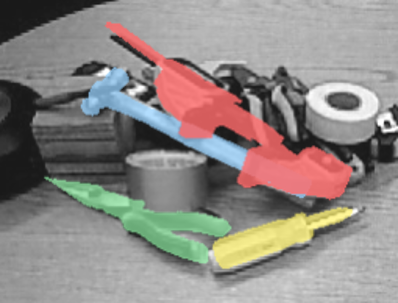}
         \caption{Localized objects in RGB image}
         \label{fig:compelling_loc_rgb}
     \end{subfigure}
     \hfill
     \begin{subfigure}[b]{0.24\textwidth}
         \centering
         \includegraphics[width=0.95\textwidth]{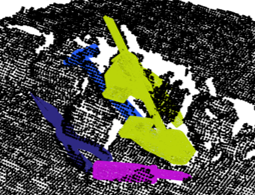}
         \caption{Localized objects in depth point cloud}
         \label{fig:compelling_loc_depth}
     \end{subfigure}
        \caption{\label{fig:qualitative_ex} Qualitative results for localization of each object in a cluttered scene using the \textit{MP + RGB-D} method. Although the segmentation map (b) is missing information for occluded parts, our iterative metho is able to recover the 6D pose of the parts (c), (d).}
\end{figure*}

\textbf{\textit{Message Passing: RGB Unary (MP+RGB): }}
To test the effect of the depth component of the unary potential, we evaluate using only the RGB component, informed by the heatmap, such that Equation \ref{eq:unary_rgbd} becomes $\phi_s(X_s, Z_s) = \phi_{s}^{\mathrm{rgb}}(X_s, Z_s^{rgb})$. We use message passing to calculate the final likelihood for each particle using the pairwise potential, as described in Equation \ref{eq:reweight_mp}. Using only the RGB image, we obtain lower accuracy on the pose estimates. The RGB image captures the position and orientation of the objects well \textit{in image space} (see Figure \ref{fig:qualitative_failures}(c)),  but is prone to falling into local minima in the axes which are not well represented by the image, namely $z$, pitch, and roll (see Figure \ref{fig:qualitative_failures}(d)).

\textbf{\textit{Message Passing: RGB Unary and ICP (MP+RGB+ICP): }}
To attempt to recover from the errors in $z$, pitch, and roll, we add an ICP step on the final estimate from \emph{MP+RGB} to align it to the masked depth image. We estimate the offset in the $z$-axis based on the depth image. Since ICP is a local refinement method which relies on an accurate estimation of the initial transform, the ICP step does not always reconcile the orientation error in pitch and roll, for which the initial transform is unknown.

\textbf{\textit{Message Passing: RGB-D Unary (MP+RGB-D): }}
We hypothesize that by including depth information in the unary potential, we can more reliably estimate the full 6D pose of the parts and make up for missing information in the 2D image. The depth term improves the estimation accuracy by discouraging all unoccluded particles from deviating from the depth image at each iteration. This performs better than \textit{MP+RGB+ICP} because the latter only attempts to align to the depth image in the final iteration, where it often has converged to a local minimum. This is the best performing method. Selected qualitative results are shown in Figures \ref{fig:qualitative_method} and \ref{fig:qualitative_ex}.

\textbf{\textit{Message Passing with Augmentation (MP+RGB-D+Aug): }}
Using the unary informed by both RGB and depth, as well as message passing, we augment the particle set at the beginning of each iteration as described in Section \ref{sec:method_particle_filter}. We use $\alpha=1.5$, with 5\% of the additional particles drawn from the unary distribution. At the first iteration,
the remaining 95\% of the particles are drawn from $q^{rand}$. The percentage of particles drawn from $q^{pair}$ is increased by 10\% every 5 iterations, and the percentage from $q^{rand}$ is decreased, up to a maximum of 90\% of particles from $q^{pair}$.

Qualitatively, we observe that the augmentation step leads to quicker convergence in some highly cluttered scenarios. On average, this method performs worse than \emph{MP+RGB-D} in some cases because it is susceptible to propagating incorrect estimates with artificially inflated pairwise scores, due to the addition of perfectly compatible pose estimates. However, these results depend on careful selection of parameters, and might be improved by further tuning.
Further analysis of the effect of the parameters on the final estimate is left to future work.

\subsection{Analysis on Tool Classes}

The results for the \textit{MP+RGB-D} and \textit{MP+RGB-D+Aug} methods for each object in the dataset are shown in Table \ref{tab:per_class_estimates}. We present the percentage of class indices which have error under 4 cm. We observed high error in the flashlight which is likely due to its symmetrical nature. The unary potential does not explicitly encode texture information, so geometrically symmetric parts can tend to flip. The clamp is among the most difficult objects to localize due to its high-dimensionality and significant self-occlusions.

\begin{table}
\renewcommand{\arraystretch}{1.3}
    \centering
    \caption{Fraction of tools with matching error less than 4 cm. All objects are evaluated with the average pairwise matching error $m$, except the screwdriver which uses the symmetric matching error, $m_{\mathrm{sym}}$.}
    \begin{tabular}{r | c c c}
    \hline
        \textbf{Class} & \emph{MP+RGB-D} & \emph{MP+RGB-D+Aug} \\
    \hline
        clamp       & 0.677 & 0.387\\
        hammer      & 0.920 & 0.400\\
        longnose pliers   & 1.000 & 0.938\\
        lineman's pliers A & 0.909 & 0.818\\
        lineman's pliers B & 0.917 & 0.833\\
        boxcutter   & 0.833 & 0.750\\
        flashlight  & 0.640 & 0.640\\
        screwdriver ($m_{\mathrm{sym}}$) & \textit{0.759} &\textit{ 0.689}\\
    \hline
    \end{tabular}
    \label{tab:per_class_estimates}
\end{table}

\subsection{Qualitative Analysis}

Selected examples of the \textit{MP+RGB-D} method are shown in Figure \ref{fig:qualitative_method}. We show a selected scene which demonstrate the effectiveness of the \textit{MP+RGB-D} method at localizing hand tools, even under partial or full occlusion of some of their parts, in Figure \ref{fig:qualitative_ex}. While the heatmap may provide little to no information for some parts, by leveraging geometric information through message passing, we are able to resolve the pose of all the visible tools in the scene.

\section{Conclusion} \label{sec:conclusion}
In this work, we present an inference technique for estimating articulated parts-based object pose in clutter. We model the part poses of each articulated object as a Markov Random Field (MRF) and perform efficient particle-based belief propagation. We use articulation constraints between parts and a novel learned likelihood function to perform message passing in the MRF. We perform a thorough analysis of our method and show that it performs well on both uncluttered and cluttered scenes. We demonstrate that the message passing step is highly beneficial in terms of enforcing geometric consistency to inform pose estimation in the high dimensional space of 6D articulated object pose.

\section*{Acknowledgements}
We would like to thank Zhen Zeng for helpful discussions and feedback, and Adrian R\"{o}fer and Zhiming Ruan for creating the 3D tool models. Zhiming Ruan also contributed to initial software development.

\bibliographystyle{IEEEtran}
\bibliography{ref.bib}

\end{document}